\documentclass[letterpaper, 10 pt, conference]{ieeeconf}  
\IEEEoverridecommandlockouts                              

\usepackage{amsmath,amsfonts,amssymb, bm}

\usepackage{cite}
\usepackage{hyperref}
\usepackage[pdftex]{graphicx}
\usepackage{epstopdf}
\usepackage{array}
\usepackage{wasysym}
\usepackage{bm}
\usepackage[per-mode=symbol]{siunitx}
\usepackage{url}
\usepackage{color}
\usepackage{epstopdf}
\usepackage[ruled]{algorithm2e}
\usepackage{multirow}
\usepackage{booktabs}
\usepackage{lineno}
\usepackage{bm}
\usepackage{mathrsfs}
\usepackage{upgreek}

\title{\LARGE \bf Integration of Riemannian Motion Policy with Whole-Body Control for Collision-Free Legged Locomotion}

\author{Daniel Marew $^{1}$, Misha Lvovsky$^{1}$, Shangqun Yu$^1$, Shotaro Sessions$^1$, and Donghyun Kim$^{1}$
\thanks{Authors are with the $^{1}$ Manning College of Information and Computer Sciences at University of Massachusetts Amherst, Amherst, MA, 140 Governors Dr, Amherst, MA 01002, USA. Corresponding Author: {\tt \small robot.dhkim@gmail.com}}%
}

\begin{document}

\maketitle
\thispagestyle{empty}
\pagestyle{empty}

\begin{abstract}
In this paper, we present a Riemannian
Motion Policy (RMP)flow-based whole-body control framework for improved dynamic legged locomotion. RMPflow is a differential geometry-inspired algorithm for fusing multiple task-space policies (RMPs) into a configuration space policy in a geometrically consistent manner. RMP-based approaches are especially suited for designing simultaneous tracking and collision avoidance behaviors and have been successfully deployed on serial manipulators. However, one caveat of RMPflow is that it is designed with fully actuated systems in mind. In this work, we, for the first time, extend it to the domain of dynamic-legged systems, which have unforgiving under-actuation and limited control input. Thorough push recovery experiments are conducted in simulation to validate the overall framework. We show that expanding the valid stepping region with an RMP-based collision-avoidance swing leg controller improves balance robustness against external disturbances by up to 53\% compared to a baseline approach using a restricted stepping region. Furthermore, a point-foot biped robot is purpose-built for experimental studies of dynamic biped locomotion. A preliminary unassisted  in-place stepping experiment is conducted to show the viability of the control framework and hardware.
\end{abstract}


\section{Introduction}
\subsection{Motivation}
In dynamic legged locomotion, the size of valid stepping region is crucial as it determines the maximum CoM velocity that can be safely regulated\cite{Pratt2006}. Previous work on dynamic legged locomotion conservatively restricted the stepping region in the lateral direction so that the robot's legs do not cross one another in order to mitigate the risk of self-collision\cite{IJRR, stabilizing, Kim2014}. However, this restriction prevents the robots from taking aggressive but stabilizing steps, thus reducing their balance stability and robustness to external disturbances. 

In this work, we aim to address this undue trade-off by deploying an RMPflow-based \cite{RMPflow_0} reactive collision-avoidance swing leg controller. The controller steers the swing foot towards the planned step location while avoiding collisions between the robot's links. This approach enables full utilization of the kinematically reachable region as a valid stepping area, thereby widening the robot's  stability margin  and improving its robustness to external disturbances.
\begin{figure}
    \centering
    \includegraphics[width=\columnwidth]{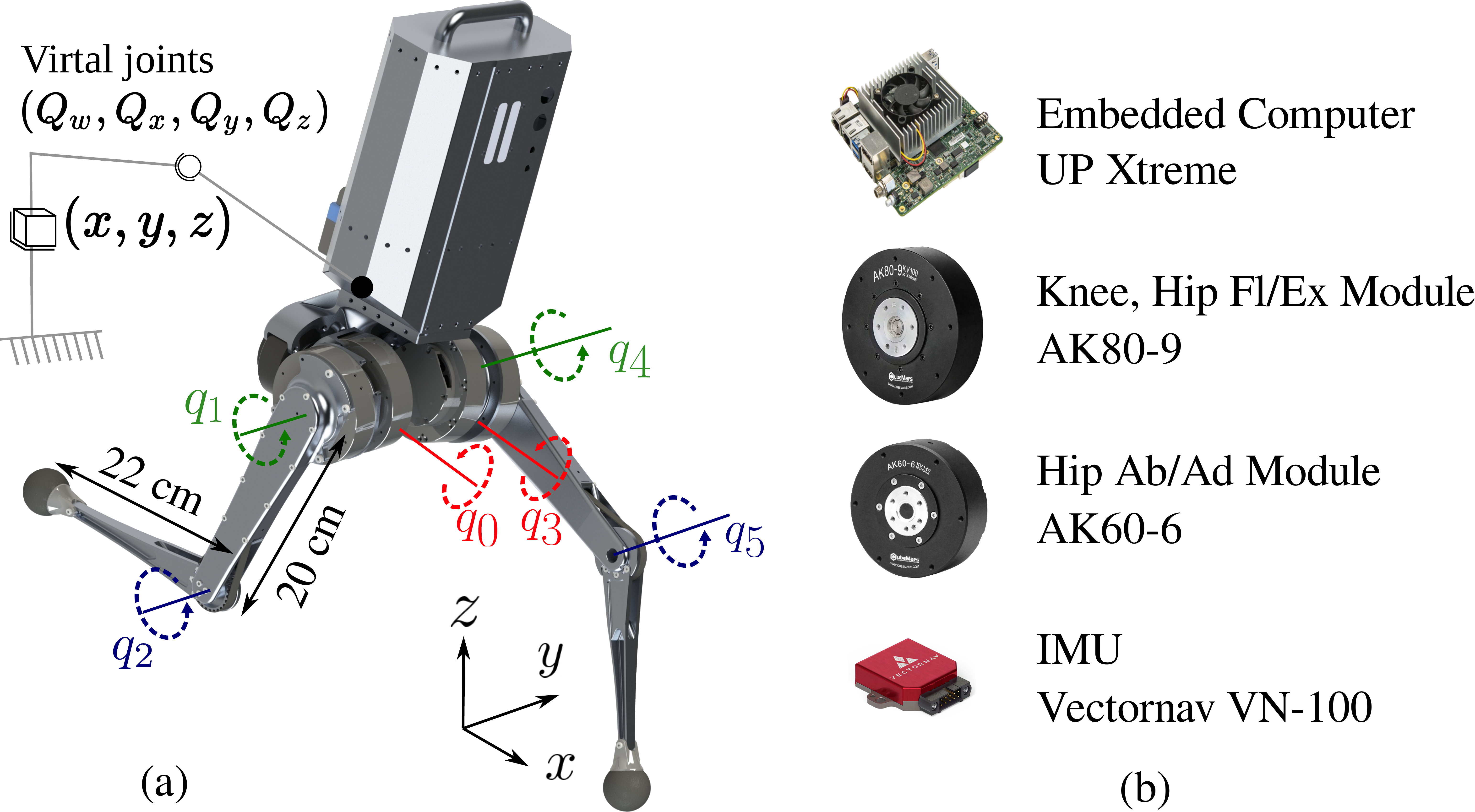}
    \caption{{\bf Point-foot biped robot, Pat.} (a) Small scale point-foot biped with 6 actuated degrees of freedom. (b) Off-the-shelf components used to build the robot. }
    \label{fig:pat_config}
    \vspace{-4mm}
\end{figure}
\subsection{Related Works}
Collision avoidance algorithms can generally be divided into two categories. The first encompasses planning-based approaches wherein long horizon collision-free trajectories are generated using either sampling-based motion planning techniques, such as Rapidly-Exploring Random Trees (RRTs) \cite{planning_rrt_1, planning_rrt_2, planning_rrt_3}, or optimization-based techniques \cite{planner_opt_1, planner_opt_2, planner_opt_3}. Although these methods are commonly employed in robotic manipulators, they tend to be computationally intensive, rendering them less suitable for real-time applications, especially those involving high-speed movements as found in dynamic legged locomotion.

The second strategy to tackle the collision avoidance problem involves the application of reactive controllers. Historically, Artificial Potential Field (APF)-based strategies have been the predominant technique for designing reactive collision avoidance behaviors. APF-based methods achieve collision avoidance by applying a virtual repulsive force to the robot to steer it away from obstacles \cite{afp_1, afp_2}. However, these techniques can pose calibration challenges, often neglect full-body dynamics, and frequently produce undesired behaviors due to local minima.

Another widely used reactive method is using a constrained optimization \cite{opt_1, cbf1, mit_collision}. In this context, the recently proposed Control Barrier Functions (CBFs) are particularly noteworthy. Specifically, Khazoom et al. derived acceleration level self-collision avoidance constraints using signed distance-based CBFs and incorporated them into a whole-body control QP formulation. However, these methods do not tend to scale efficiently as the number of considered collision elements increases. Each additional collision element introduces a new constraint to the optimization problem, potentially rendering these methods impractical for fast and real-time applications. Moreover, conflicts between these constraints can sometimes render the optimization problem infeasible \cite{mit_collision}.


RMPflow \cite{RMPflow_0} is a promising alternative to these approaches because it scales well with the number of collision tasks requiring a limited amount of additional computation power, making it suitable for fast and real-time operations. Moreover, it mitigates local-minima-induced undesired behaviors of APF-based methods by specifying the behavior with an additional highly non-linear priority Riemannian metric that stretches the space in the direction of obstacles\cite{RMPflow_1}. This priority metric heavily penalizes motion in the direction of obstacles while being indifferent to motion in the orthogonal direction\cite{RMPflow_0}. In addition, because the priority metric can be velocity dependent, it can be designed so that the robot ignores nearby obstacles it is moving away from seamlessly\cite{RMPflow_1}. This behavior is crucial for high-speed operations and cannot be achieved with APF methods that rely solely on position-dependent repulsive forces.

In the RMPflow framework\cite{RMPflow_0,RMPflow_1}, each desired task-space behavior is specified by a local motion policy called a Riemannian motion policy (RMP), which is composed on the task space manifold. An RMP consists of an acceleration policy and an associated Riemannian metric, also known as a state-dependent inertia metric, that encodes information about the task space's geometry and the relative priority of the task. The RMPflow algorithm fuses these local RMPs into a single configuration-space ($\mathcal{C}$-space) RMP using an operator called the pullback operator.

One caveat of RMPflow is its implicit assumption that the $\mathcal{C}$-space acceleration policy obtained through the pullback operations is always realizable by the robot. This assumption is often not satisfied when it comes to underactuated systems like legged robots. To-date, the framework has been primarily applied to fully actuated systems such as serial manipulators\cite{RMPflow_0, RMPflow_1, RMPflow_3, RMPflow_4, rmp_flow_5}, with one notable exception\cite{extending}. Wingo et al. \cite{extending} proposed an extension of RMPflow for a class of underactuated wheeled inverted pendulum (WIP) robots.
An actuated joint directly controls the floating base of the class of WIP robots considered in \cite{extending, extending_2}. The underactuation of this class of robots only emanates from the fact that this joint shares the same control torque as the wheelbase. 
This kind of underactuation lends itself to the type of dynamics separation scheme the RMPflow formulation in \cite{extending} relies on. This underactuation,  however, differs from the underactuation of legged systems whose floating base can not be directly controlled \cite{extending_2}. Thus the RMPflow framework developed in \cite{extending} can not be directly applied to legged systems.

In this paper, we address this gap by integrating RMPflow with a traditional null space projection-based whole-body controller formulation so that tasks specified in terms of RMPs can be realized on legged systems. In our approach, the RMPs are executed in a way that is consistent with robot's contact-constrained dynamics while not compromising the tracking performance of higher priority tasks such as floating base tasks. Specifically, we formulate a constrained weighted least-squares problem inspired by the RMPflow formulations in \cite{RMPflow_0} and \cite{extending} whose optimal solution, a $\mathcal{C}$-space acceleration command,  will try to  realize RMPs as faithfully as possible, giving priority according to the assigned metric. Moreover, the solution is by construction guaranteed not to violate contact constraints and constraints derived from higher priority tasks.



Point-foot bipeds are relatively simple legged systems, but since they are severely underactuated and highly unstable, they are extremely challenging to control, making them an ideal experimental platform for testing the limits of dynamic biped locomotion. To that end, this paper presents a new small point-foot biped named Pat designed and built to experimentally validate the proposed whole-body framework. 


In summary, the key contributions of our study include the following:
\begin{enumerate}
    \item We present a new formulation for integrating RMPflow with a null-space projection-based whole-body controller, enabling its application in legged robots.
    \item Our extensive simulation experiments highlight that our proposed collision-avoidance swing-leg controller, developed based on the aforementioned formulation, significantly enhances the robustness of a point-foot biped robot against external disturbances. In addition, we apply and test high-speed self-collision avoidance on a quadruped robot, and provide a comparative analysis of our results against APF and CBF based methods.
    \item We designed a low cost point-foot biped robot for the experimental study of dynamic biped locomotion. The performance of our proposed controller and the viability of the robot's hardware is validated through successful demonstrations of unassisted, in-place walking.
\end{enumerate}

\section{Integration of Riemannian Motion Policy and Whole-Body Control}
This section details the proposed approach for integrating RMPflow with the traditional null-space projection-based whole-body controller formulation. We use the following general equations of motion of legged robots
\begin{equation} \label{eq:full_dyn}
    \bm{A}
\ddot{\mathbf{q}}+ \mathbf{b}+\mathbf{g} = 
    \bm{S}_a^\top\bm{\tau}
    +\bm{J}_c^{\top}\mathbf{f}_r,
\end{equation}
where $\bm{A} \in \mathbb{R}^{n+6\times n+6}$, $\mathbf{b}\in \mathbb{R}^{n+6}$, and $\mathbf{g}\in \mathbb{R}^{n+6}$, are the generalized mass matrix, coriolis force, and gravitation forces, respectively.
$\bm{S}_a \in \mathbb{R}^{n+6\times n+6}$ is the actuated joint
selection matrix. $\bm{\tau} \in \mathbb{R}^{n+6}$, $\mathbf{f}_r \in \mathbb{R}^{3\cdot n_c}$, and $\bm{J}_c \in \mathbb{R}^{3\cdot n_c\times n+6}$ are joint torque, augmented reaction force and contact Jacobian, respectively. $\ddot{\mathbf{q}}\in \mathbb{R}^{6+n}$ is the configuration space acceleration where $n$ the number of actuated degrees of freedom and $n_c$ is the number of contact points.



\subsection{Review of RMPflow}
In this section, we briefly introduce the RMPflow computational framework first proposed in \cite{RMPflow_0}. RMP composed on an m-dimensional manifold $\mathcal{M}$ is characterized, in its canonical form, by
the tuple $(\mathbf{a}, \bm{M})^\mathcal{M}$, where $\mathbf{a}: \mathbb{R}^{m}\times\mathbb{R}^{m}\rightarrow\mathbb{R}^{m}$ is an acceleration policy and $\bm{M}: \mathbb{R}^{m}\times\mathbb{R}^{m}\rightarrow\mathbb{R}^{m \times m}_{+}$ is a state-dependent Riemannian metric. This tuple can be written in its natural form, $(\mathbf{f}, \bm{M})^\mathcal{M}$, by using $\mathbf{f} = \bm{M} \mathbf{a}$ where $\mathbf{f}$ can be considered as a virtual force. 

RMPflow uses a tree-like data structure called RMP-tree to efficiently encode the hierarchical relationship between RMPs. 
In general, operational space tasks specified in terms of RMPs are placed at the leaves of the RMP tree, and the root-node RMP represents the configuration space policy. RMPflow uses three operators that consitute the RMP-algebra. These are; \textbf{pushforward}, \textbf{pullback}, and \textbf{resolve}. In the first stage of the RMPflow algorithm, the \textbf{pushforward} operator is used to propagate state (position and velocity) information from the root, $\mathcal{C}$-space, to the leaves of the RMP tree, task space. In the case of typical whole-body control formulation, this is similar to updating the position and velocity of task space based on the state of the configuration space using forward kinematics and the Jacobian map, respectively. 

In the subsequent stage of RMPflow, starting at the RMP tree leaves, the pullback operator is recursively applied using (\ref{eq:rmp_pullback}) to obtain the combined inertia metric and virtual force of the parent RMPs and eventually that of the $\mathcal{C}$-space RMP. 
\begin{align}
    \label{eq:rmp_pullback}
    \mathbf{f}' = \sum_{i=1}^{K} \bm{J}_i^{\top}\left ( \mathbf{f}_{i}  - \bm{M}_i\dot{\bm{J}}_i\dot{\mathbf{x}}\right), & \hspace{0.2cm} \bm{M}' = \sum_{i=1}^{K} \bm{J}_i^{\top}\bm{M}_i\bm{J}_i, 
\end{align}
where $(\mathbf{f}_i, \bm{M}_i)^{\mathcal{N}_i}$ denotes the $i^{\rm th}$ child RMP, $\bm{J}_i$ represents the associated Jacobian matrix, $\dot{\mathbf{x}}_i$ is the task space velocity, and $K$ is the total number of child RMPs. 
Lastly, \textbf{resolve}
 is used to transform the computed natural form RMP, $(\mathbf{f},\bm{M})^{\mathcal{M}}$ to its canonical form $(\mathbf{a},\bm{M})^{\mathcal{M}}$ where $\mathbf{a} = \mathbf{M}^{\dagger}\mathbf{f}$ and ${\dagger}$ denotes Moore-Penrose inverse.

\subsection{Review of null-space projection based task prioritization}
\label{sec:nullspace}
We employ the dynamically consistent null space projection technique to impose a strict hierarchy between tasks. It can be described with the following recursion rule from~\cite{Kim:2018iq}.
\begin{equation}
\begin{split}
\ddot{\mathbf{q}}^{\rm cmd}_{i} &= \ddot{\mathbf{q}}^{\rm cmd}_{i-1} + \overline{\bm{J}_{i|pre}^{\rm dyn}}\left( \ddot{\mathbf{x}}_{i}^{\rm cmd} - \dot{\bm{J}}_i\dot{\mathbf{q}} - \bm{J}_{i} \ddot{\mathbf{q}}_{i-1}^{\rm cmd} \right), \\[2mm]
\label{eq:qddot_cmd}
 \end{split}
\end{equation}
\vspace{-0.5cm}
where
\begin{align}
	&\bm{J}_{i|pre} = \bm{J}_i \bm{N}_{i-1}, \\[2mm]
\begin{split}
    &\bm{N}_{i-1} = \bm{N}_{0}\bm{N}_{1|0} \cdots \bm{N}_{i-1|i-2},\\
    &\bm{N}_{i|i-1} = \bm{I} - \overline{\bm{J}_{i|i-1}^{\rm dyn}}\bm{J}_{i|i-1},\\
    &\bm{N}_{0} = \bm{I} - \overline{\bm{J}_{c}^{\rm dyn}}\bm{J}_c,
    \end{split}
\end{align}
Here, $i\geq 1$, and 
\begin{equation}
    \begin{split}
        \label{eq:contact}
        \ddot{\mathbf{q}}_0^{\rm cmd} &= \overline{\bm{J}_c^{\rm dyn}}(-\bm{J}_c\dot{\mathbf{q}}).
    \end{split}
\end{equation}
$\ddot{\mathbf{x}}^{\rm cmd}_i$ is the acceleration policy of $i$-th task defined by the PD controller.
\begin{equation}
    \label{eq:attractor_rmp}
    \ddot{\mathbf{x}}^{\rm cmd}_i(\mathbf{x}, \dot{\mathbf{x}}) = \ddot{\mathbf{x}}^{\rm des} + \bm{K}_p \left(\mathbf{x}^{\rm des}_i - \mathbf{x}_i\right) + \bm{K}_d\left(\dot{\mathbf{x}}^{\rm des} - \dot{\mathbf{x}}\right),
\end{equation} 
where $\bm{K}_p$ and $\bm{K}_d$ are position and velocity feedback gains, respectively. $\bm{J}_{i|pre}$ is the projection of the $i$-th task Jacobian into the null space of the prior tasks. $\bm{J}_c$ is a contact Jacobian and the dynamically consistent pseudo-inverse is denoted by an overline and superscript `dyn' as in~\cite{wbic} and  defined as
\begin{equation}
    \overline{\bm{J}^{\rm dyn}} = \bm{A}^{-1}\bm{J}^{\top}\bm{\Lambda}.
\end{equation}
Here, $\bm{\Lambda}$ is the operational mass matrix given by 
\begin{equation}
    \label{eq:osc_mm}
    \bm{\Lambda} = \left( \bm{J} \bm{A}^{-1} \bm{J}^{\top} \right)^{-1}.
\end{equation}
We would like to reiterate that the contact task is always given the highest priority (\ref{eq:contact}) and all tasks including RMP-based tasks in \ref{swing_leg} are executed in the null-space of this task to prevent contact constraint violations.

\subsection{Integrating RMP with prioritized task execution}
In this section, we will provide a modification of the pullback operation at the root of the RMP-Tree, which computes the final $\mathcal{C}$-space acceleration command. The modification allows us to realize tasks specified in terms of RMPs in a way that does not interfere with the execution of higher priority tasks. To do so, we define the final $\mathcal{C}$-space acceleration command in the following way.
\begin{equation}
\label{q_cmd}
\ddot{\mathbf{q}}^{\rm cmd} = \ddot{\mathbf{q}}^{\rm cmd}_{k} + \bm{N}_{k}\ddot{\mathbf{q}}_{\rm rmp}
\end{equation}
where $\ddot{\mathbf{q}}^{\rm cmd}_{k}$ and $\bm{N_k}$ are the acceleration command and null space projection matrix derived from the first $k$ higher priority tasks, including the contact constraint, using successive null space projections described in \ref{sec:nullspace}. $\ddot{\mathbf{q}}_{\rm rmp}$ is any arbitrary $\mathcal{C}$-space acceleration that can be used to realize RMPs. We obtain the optimal $\mathcal{C}$-space acceleration command by solving the following constrained weighted least-squares problem.
\begin{equation}
\label{eq:RMP_QP_con}
\begin{split}
 \min_{\ddot{\mathbf{q}}^{\rm cmd}} &\sum_{i=0}^{n} || \bm{J}_i \ddot{\mathbf{q}}^{\rm cmd} + \dot{\bm{J}}_i\dot{\mathbf{q}} - \ddot{\mathbf{x}}_i||^2_{\mathbf{M}_i} \\
\text{s.t.} & \\
&\ddot{\mathbf{q}}^{\rm cmd} = \ddot{\mathbf{q}}^{\rm cmd}_{k} + \bm{N}_{k}\ddot{\mathbf{q}}_{\rm rmp}
\end{split}
\end{equation}
Where $(\ddot{\mathbf{x}}_i, \mathbf{M}_i)^{\mathcal{N}_i}$ is the $i^{\rm th}$ child RMP of the root RMP and $\bm{J}_i$ is the Jacobian that maps $\mathcal{C}$-space velocities to the $i^\text{th}$ child nodes' task space. The optimal solution to (\ref{eq:RMP_QP_con}) will try to realize all the tasks specified in the RMP-tree as faithfully as possible, giving priority according to the assigned metric $\mathbf{M}_i$. By construction it is guaranteed not to violate constraints obtained from the higher priority tasks and is consistent with the contact constrained dynamics of the robot.

The constrained least-squares problem in (\ref{eq:RMP_QP_con}) can be reformulated into an unconstrained least-squares problem in (\ref{eq:RMP_QP}) by moving the constraint to the cost.
 \begin{equation}\label{eq:RMP_QP}
      \ddot{\mathbf{q}}_{\rm rmp}^* = \arg \min_{\ddot{\mathbf{q}}_{\rm rmp}}\sum_{i=0}^{n} || \bm{J}_i\bm{N}_{k}{\ddot{\mathbf{q}}}_{\rm rmp} - \hat{\ddot{\mathbf{x}}}_i||^2_{\mathbf{M}_i}
\end{equation}
where  $\hat{\ddot{\mathbf{x}}} =  \ddot{\mathbf{x}}_i - \dot{\bm{J}}_i\dot{\mathbf{q}} - \bm{J}_i\ddot{\mathbf{q}}_{k}$. The analytical solution to (\ref{eq:RMP_QP}) yields the modified pullback operation given by (\ref{eq:m_rmp})-(\ref{eq:f_rmp}).


%
\vspace{-5mm}
    \begin{align}
        \label{eq:m_rmp}
        \mathbf{M}_{\rm rmp} &= \bm{N}_k^{\top}\left(\sum_{i=0}^{n} \bm{J}_i^{\top}\mathbf{M}_i\bm{J}_i\right)\bm{N}_k \\[2mm]  
    \label{eq:f_rmp}
    \mathbf{f}_{\rm rmp} &= \bm{N}_k^{\top} \left(\sum_{i=0}^{n} \bm{J}_i^{\top}\left(\mathbf{f}_i - \mathbf{M}_i\dot{\bm{J}}_i\dot{\mathbf{q}} \right)- \bm{J}_i^{\top}\mathbf{M}_i\bm{J}_i\ddot{\mathbf{q}}_{k}\right)
    \end{align}
where $\mathbf{M}_{\rm rmp}$ and $\mathbf{f}_{\rm rmp}$ are the the projected virtual inertia metric and virtual force computed through the modified pullback operations in (\ref{eq:m_rmp}) and (\ref{eq:f_rmp}) respectively. Note the term inside the bracket in (\ref{eq:m_rmp}) and the first term in (\ref{eq:f_rmp}) are the regular RMPflow pullback operations in (\ref{eq:rmp_pullback}). Thus $\mathbf{M}_{\rm rmp}$, $\mathbf{f}_{\rm rmp}$  can be obtained from the $\mathcal{C}$-space RMP $(\mathbf{M}, \mathbf{f})^{\mathcal{M}}$ (\ref{eq:RMP_final}). 
\begin{equation}\label{eq:RMP_final}
    \begin{split}
    \hspace{2mm}
        \mathbf{M}_{\rm rmp} = \bm{N}_k^{\top}\mathbf{M}\bm{N}_k,  & \hspace{2mm}
        \mathbf{f}_{\rm rmp} = \bm{N}_k^{\top}\left(\mathbf{f} -  \mathbf{M}\ddot{\mathbf{q}}^{\rm cmd}_{k}\right)
    \end{split}
\end{equation}
The final acceleration command is then obtained by
\begin{equation}
\label{q_cmd_final}
\ddot{\mathbf{q}}^{\rm cmd} = \ddot{\mathbf{q}}^{\rm cmd}_{k} + \bm{N}_{k}\mathbf{M}_{\rm rmp}^{\dagger} \mathbf{f}_{\rm rmp} 
\end{equation}

\subsection{Whole-body control (WBC)}
\label{wbc}
In this work, we execute the acceleration command from (\ref{q_cmd_final}) using a convex MPC \cite{cMPC} and whole body impulsive control (MPC
+ WBIC) based controller proposed in \cite{wbic}. In \cite{wbic}, an optimal reaction force profile is computed using a convex MPC based on a simplified single rigid body model of the robot. This reaction force is then tracked along side acceleration commands while considering the full-body dynamics of the robot by solving the following quadratic program (QP) (\ref{eq:qp_cost}).  
Once the optimal $\mathbf{f}_r$ and $\ddot{\mathbf{q}}$ are obtained by solving the QP in (\ref{eq:qp_cost}), inverse dynamics is used to compute the torque command. 
\begin{equation} \label{eq:qp_cost}
\min_{\bm{\delta}_{\mathbf{f}_r}, \bm{\delta}_{f}}\quad \bm{\delta}_{\mathbf{f}_r}^{\top} \bm{Q}_1 \bm{\delta}_{\mathbf{f}_r} + \bm{\delta}_{f}^{\top}\bm{Q}_2\bm{\delta}_{f}\vspace{0.7mm} \\
 \vspace{-4mm}
\end{equation}
\begin{align*}
\text{s.t.} & \\
\tag{floating base dyn.}
&\bm{S}_f 
\left(
\bm{A} \ddot{\mathbf{q}} + \mathbf{b} + \mathbf{g}
\right) 
= \bm{S}_f \bm{J}_{c}^{\top} \mathbf{f}_r \\
\tag{acceleration}
&\ddot{\mathbf{q}} = \ddot{\mathbf{q}}^{\rm cmd} +  \begin{bmatrix}
\bm{\delta}_{f} \\ \mathbf{0}_{n_j}
\end{bmatrix} \\
\tag{reaction forces}
&\mathbf{f}_r 
= \mathbf{f}_r^{\rm MPC} + \bm{\delta}_{\mathbf{f}_r}\\
\tag{contact force constraints}
 & \bm{W} \mathbf{f}_r \geq \mathbf{0},
\end{align*}
where $\mathbf{f}_r^{\rm MPC}$ is the reaction force command computed by the MPC, and $\bm{S}_f$ is the floating base selection matrix. $\bm{J}_c$ and $\bm{W}$ are the augmented contact Jacobian and contact constraint matrix respectively. $\bm{\delta}_{f}$ and $\bm{\delta}_{\mathbf{f}_r}$ are relaxation variables for the floating base acceleration and reaction forces.


\section{Point-foot Biped Locomotion}
\begin{figure*}[ht]
\centering
\includegraphics[width=2\columnwidth]{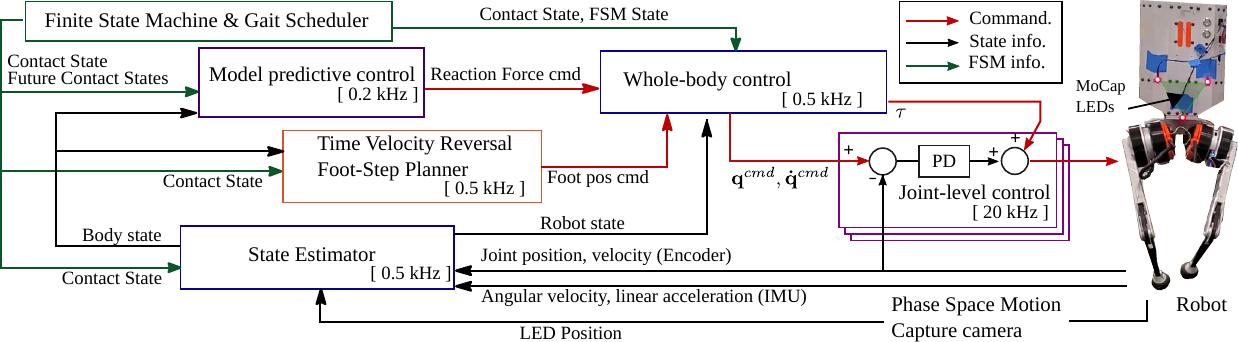}
\caption{{\bf Overall Control Framework.} A time-scripted finite state machine is used to drive the system. WBC takes reference reaction force from MPC, foot trajectory from the TVR planner, and constant body posture command. Then WBC outputs joint position, velocity and toque commands which are then executed by the joint-level controllers. The body velocity is estimated by MoCap LED position data and used for foot step planning. The WBC in this figure includes the prioritized null-space projection, RMPflow and the QP computation.}
\label{fig:ctrl_framework}
\vspace{-4mm}
\end{figure*}

This section details the design of the experimental platform and the components of the proposed control framework summarized in Fig.\ref{fig:ctrl_framework}.

\subsection{Experimental platform} 
This section details the components used in our new point-foot bipedal robot Pat depicted in Fig. \ref{fig:pat_config}. 
Pat is a 70cm tall 5.4kg 6-DoF point-foot biped designed to be used as a small-scale, low-cost experimental platform that can be quickly built and tested without requiring significant maintenance effort. Thus, it primarily employs off-the-shelf components. Pat's legs are a version of the MIT-Mini-Cheetah \cite{mini-cheetah} legs modified to work with off-the-shelf actuators. We use two types of torque-controlled electric actuators from T-motor to drive the robot's six actuated degrees of freedom. These are the AK80-9 and the AK60-6 actuators \cite{tmotor}, which consist of thin, large-diameter out-runner motors, a motor controller based on the open source Mini-Cheetah motor controller\cite{mini-cheetah}, and a planetary gear reducer embedded into the stator of the motor. The gear reductions for the actuator modules are 9:1 and 6:1, respectively. The hip abduction and flexion axes on each leg are driven directly by the AK60-6 and AK80-9 actuators, respectively, and each knee flexion joint is driven by an AK80-9 actuator connected to a timing belt. The timing belt affords the actuator used on the knee an additional reduction factor of $\frac{14}{9}\approx1.5$. 
An onboard UP-Xtreme computer with Intel Core i3-8145UE  processor running  CONFIG\_PREEMPT\_RT patched Ubuntu 18.04 is used to control the robot. The computer sends commands and receives sensory data to and from the actuators via CAN 2.0 communication protocol at 500 hz.
\subsection{Gait control}
Each gait cycle has two swing phases and two brief dual support phases. These brief dual support phases are required for point-foot bipeds to perform yaw control. In addition to the dual support and swing phases, there are four transition phases that we use to facilitate a smoother contact transition \cite{IJRR}. In particular, during these transition phases, slowly changing upper and lower bounds are introduced in the reaction force computation in (\ref{eq:qp_cost}) to avoid discontinuous reaction force commands, which can cause jerky motion.

We use the time-to-velocity reversal (TVR) planner~\cite{IJRR} to determine the upcoming footstep location. The planner is called once in the middle of each swing phase. In addition, TVR is used to generate future reference foot-step locations for the MPC. The swing leg is steered towards chosen footstep location following a minimum jerk trajectory.

\subsection{Self-collision avoidance}\label{swing_leg}
In this work, an RMPflow-based reactive collision-avoidance swing leg controller is used to steer the swing foot towards the planned step location while avoiding collisions between the robot's links. This swing-leg controller allows the robot to fully use its kinematically reachable region as a valid stepping area, thereby improving its robustness to external disturbance. We use two types of RMPs to accomplish this swing-leg behavior: an \textbf{attractor-RMP} and a set of \textbf{collision-avoidance-RMPs}. 

The \textbf{attractor-RMP} is used to move the foot towards the planned footstep location and is specified by the PD controller acceleration policy in (\ref{eq:attractor_rmp}) and the operational space inertia metric in (\ref{eq:osc_mm}). Note that the attractor RMP used in this work is not designed to meet the definition of a geometric dynamical system (GDS) required by \cite{RMPflow_1} to guarantee RMPflow's stability in the Lyapunov-sense. But, similar to Q-Function-based-RMPs described in\cite{RMPflow_0}, it can be considered a part of a broader class of RMPs that do not meet this criterion but still have practical use.

Each \textbf{collision-avoidance-RMP} is tasked with avoiding collision between a pair of control and collision points that lie on the swing leg and stance leg, respectively. For computational efficiency, we approximate the geometry of each leg's links with a capsule and reduce the link-to-link collision avoidance problem to the more simplified problem of avoiding collision between the two closest points on the capsules called the witness points. Thus, only one collision RMP is required per link pair. We use the algorithm proposed in \cite{min_distance} to compute the position of the witness points. The collision avoidance RMPs used in this work are defined on the one-dimensional Euclidean distance space and are specified by the repulsive acceleration policy and metric pair in Eq. (\ref{eq:collision_policy}) and (\ref{eq:collison_metric}), respectively \cite{RMPflow_1}.
\begin{equation}\label{eq:collision_policy}
    \ddot{x}(x,\dot{x}) = k_p\exp{(-x/l_p)}-k_d\frac{\sigma(\dot{x})\dot{x}}{x/l_d+\epsilon_d}
\end{equation}
\begin{equation}\label{eq:collison_metric}
    m(x,\dot{x}) = \sigma(\dot{x})g(x)\frac{\mu}{x/l_m + \epsilon_m}
\end{equation}
where $\sigma(\dot{x}) = 1- \frac{1}{1+\exp{(-\dot{x}/v_d})}$ and 
\begin{equation}
g(x)=
    \begin{cases}
        x^2/r^2-2x/r +1, & x \le r\\
        0, &  x > r
    \end{cases}
\end{equation}
where $x$ is the Euclidean distance between the capsule witness points and $\dot{x}$ is its rate of change. $k_p[m/s^2]$ and $k_d[s^{-1}]$ are repulsion and damping gains. The parameter $\mu$ is used to specify the priority of the collision RMP relative to other RMPs and $r[m]$ is used to control at what distance the collision RMP is disabled. $l_p[m]$, $l_m[m]$ and $v_d[m/s]$ are scaling parameters. $\epsilon_m$ and $\epsilon_d$ are offset parameters. 
\begin{figure*}[ht]
\centering
\includegraphics[width=2\columnwidth]{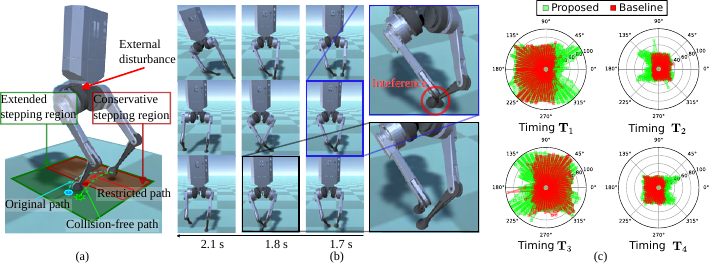}
\caption{{\bf Push recovery simulation results}  (a) The green and red boxes represent the valid stepping region of the proposed strategy and baseline strategies, respectively. The red and green dashed lines represent the foot trajectory under the baseline and proposed strategy in response to the external disturbance. 
(b) Snapshots of the robot trajectory under three scenarios. First row (baseline strategy), second row: leg crossing movement without collision avoidance, bottom row (proposed strategy): safe leg crossing movement with collision avoidance.
(c) polar coordinate representation of the the applied disturbance forces. Successful outcomes are shown for the proposed (green) and baseline (red) strategies at four different disturbance timings. $\mathbf{T}_1$ and $\mathbf{T}_3$ correspond to the 
double stance phase and $\mathbf{T}_2$ and $\mathbf{T}_4$ correspond to the right swing and left swing phases. Note that failed trials are not represented in these plots.
}
\label{fig:push_rec_snap}
\end{figure*}

\begin{figure}
    \centering
    \includegraphics[width=1\columnwidth]{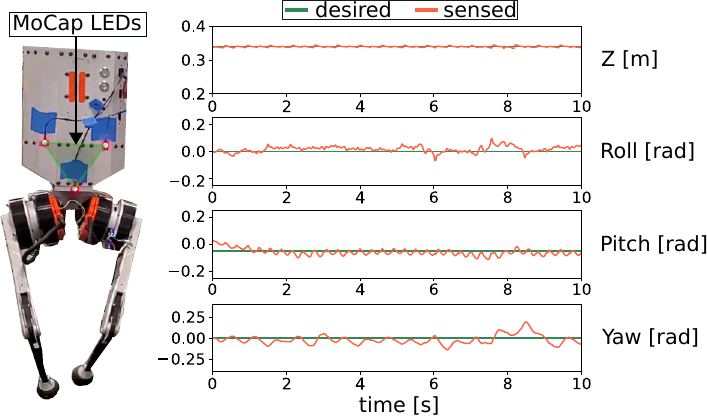}
    \caption{{\bf Hardware experiment result}. Floating-base reference tracking performance during unassisted in-place walking test.}
    \label{fig:hw_result}
    \vspace{-4mm}
\end{figure}
\subsection{State estimation}
Together with the robot's kinematic model, joint encoder and IMU data are used to estimate the floating base and CoM states. The estimated floating base position and velocity are used for feedback control. However, the CoM velocity estimate is too noisy for foot-step planning. Thus, in this paper, similar to \cite{IJRR, stabilizing}, an external MoCap system is used to estimate the base velocity based on an LED attached to the robot's body which is used as an approximate estimate of the CoM velocity. We carefully chose the base-frame position so that this approximation is valid.

\section{EXPERIMENTAL RESULTS AND ASSESSMENT}
\subsection{Robustness analysis in dynamics simulation}
\label{sim_test}
In this section, we validate the proposed locomotion controller in a kinodynamically faithful simulation of our point-foot biped robot Pat. For these simulations, we employ the multi-DoF rigid-body dynamics simulator from MIT's Biomimetic Robotics Lab, which notably incorporates actuator dynamics\cite{mit_humanoid}. The controller's task is to stabilize this highly underactuated and unstable robot while avoiding collision between the robot's links, even under external disturbance. In all experiments we define the following task hierarchy in descending order of priority: contact, body orientation, body position, RMP swing leg. The MPC horizon is set to one gait period, which is 600 ms.
\subsubsection{Experimental setup}
 In this work, we only consider the relevant body-segment collision pairs to reduce computation time. Because of Pat's morphology, the risk of collision almost exclusively exists between the two lower limbs while it is standing. Thus in all experiments, we only consider this collision pair. We approximate the geometry of the robot's lower limbs with a capsule of radius 15mm.
  The combined task projection, RMPflow, and whole body QP computations take less than 0.2 ms on an Intel i9-12900K CPU.
  
\subsubsection{Push Disturbance Resistance Test}
In order to test the effectiveness of the proposed controller, we run extensive push disturbance experiments. In particular, disturbance forces ranging between 10-100N that lie on the transverse plane are applied on the robot's base for 20ms at four instances—two swing phases and two double stance phases . Failure is reported if the base of the robot is below 25 cm or collision is detected between the two limbs during the subsequent 3 seconds after the disturbance is applied. 

Our proposed strategy allows the robot to safely perform leg crossing movements by employing the self-collision-avoidance controller described in \ref{swing_leg}. Thus, it takes advantage of the entire kinematically reachable region to recover from the disturbance. We benchmark the proposed strategy with a baseline conservative foot placement strategy that artificially restricts the lateral stepping region to avoid leg crossing movements used in \cite{Kim2014, Kim:2018iq, IJRR}.
Fig. \ref{fig:push_rec_snap} shows the results of simulation experiment. Fig. \ref{fig:push_rec_snap}(a) shows the stepping regions of the proposed and baseline strategies, which are the extended and conservative stepping regions, respectively. 
Fig. \ref{fig:push_rec_snap}(b) depicts the snapshots of the trajectories of the robot state in reaction to a lateral disturbance force $\mathbf{F}_y=-50$N. The first row depicts the robot state trajectory when the baseline strategy is used. Because it could not take the necessary stabilizing step, the robot could not recover from the disturbance and eventually fell. In the second row, the robot is shown performing leg crossing movements to recover from the disturbance. However, as is highlighted in the figure, collision instances were recorded along the way. The last row shows the robot under the proposed strategy. Under this strategy, the robot performed the leg crossing movement safely with the help of a collision avoidance controller and was able to reject the disturbance successfully.
Fig.\ref{fig:push_rec_snap}(c) and Table \ref{tab:success_rate} summarize the results of 10,000 experiments. Fig. \ref{fig:push_rec_snap}(c) shows that the proposed strategy can recover from a larger range of disturbance forces by utilizing its extended stepping region in both double stance ($\mathbf{T}_1$ and $\mathbf{T}_3$) and swing phases ($\mathbf{T}_2$ and $\mathbf{T}_4$). Table \ref{tab:success_rate} shows the success rate (SR) and conditional ratios $\eta_{p|b}$ and $\eta_{b|p}$. Where $\eta_{p|b}$ is the ratio of the number of successes of both strategies with the number of success of the baseline and $\eta_{b|p}$ is the ratio of the number of successes in both strategies with the number of success with the proposed strategy.
In more than 90\% of cases, the proposed controller can recover from disturbances the baseline strategy recovered from, but the baseline can only recover from less than 70\% of the cases the proposed strategy recovered from. Moreover, comparing their success rates shows up to 53\% improvement from the baseline strategy. 

\begin{table}
    \centering
    \setlength\tabcolsep{2.5pt}
    \begin{tabular}{c|c|c|c|c|c}
    \hline
         &  $\eta_{b|p}$ & $\eta_{p|b}$ & Baseline SR & Proposed SR & Improvement\\
    \hline
    $\mathbf{T_1}$ & 68.06\% & 93.55\% & 47.76\%& \textbf{65.64}\% & 37.43\%\\
    $\mathbf{T_2}$ & 60.56\% & 93.02\% & 18.36\%& \textbf{28.2}\% & 53.59\%\\
    $\mathbf{T_3}$ & 68.78\% & 90.87\% & 49.56\%& \textbf{65.48}\%& 32.12\%\\
    $\mathbf{T_4}$ & 64.46\% & 91.61\% & 19.56\%& \textbf{27.8}\%& 42.12\%\\
    \hline
    \end{tabular}
    \caption{Success ratio and rate comparison.}
    \label{tab:success_rate}
    \vspace{-6mm}
\end{table}

\subsection{High-speed self-collision avoidance}
\begin{figure}
    \centering
    \includegraphics[width=0.5\textwidth]{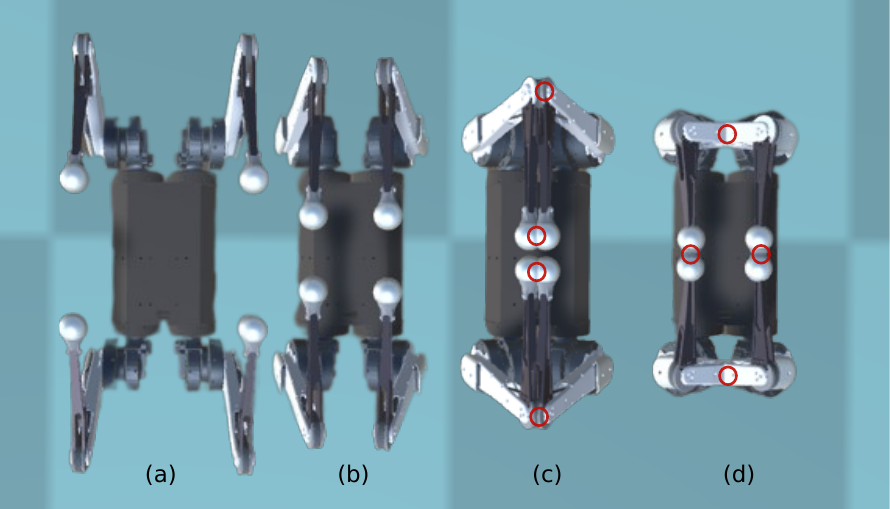}
\caption{High-speed self-collision avoidance test conducted with the MIT Mini-Cheetah. This figure presents a top-down view of the Mini-Cheetah robot, inverted and performing a leg-crossing maneuver. The cycle of movement, from (a) through (d) and back, is executed repetitively. Illustrated here is the scenario in which the collision avoidance feature is deactivated. Red circles highlight the specific locations of collision points.}
    \label{fig:mc_exp}
\end{figure}

\begin{figure}
    \includegraphics[width=0.5\textwidth]{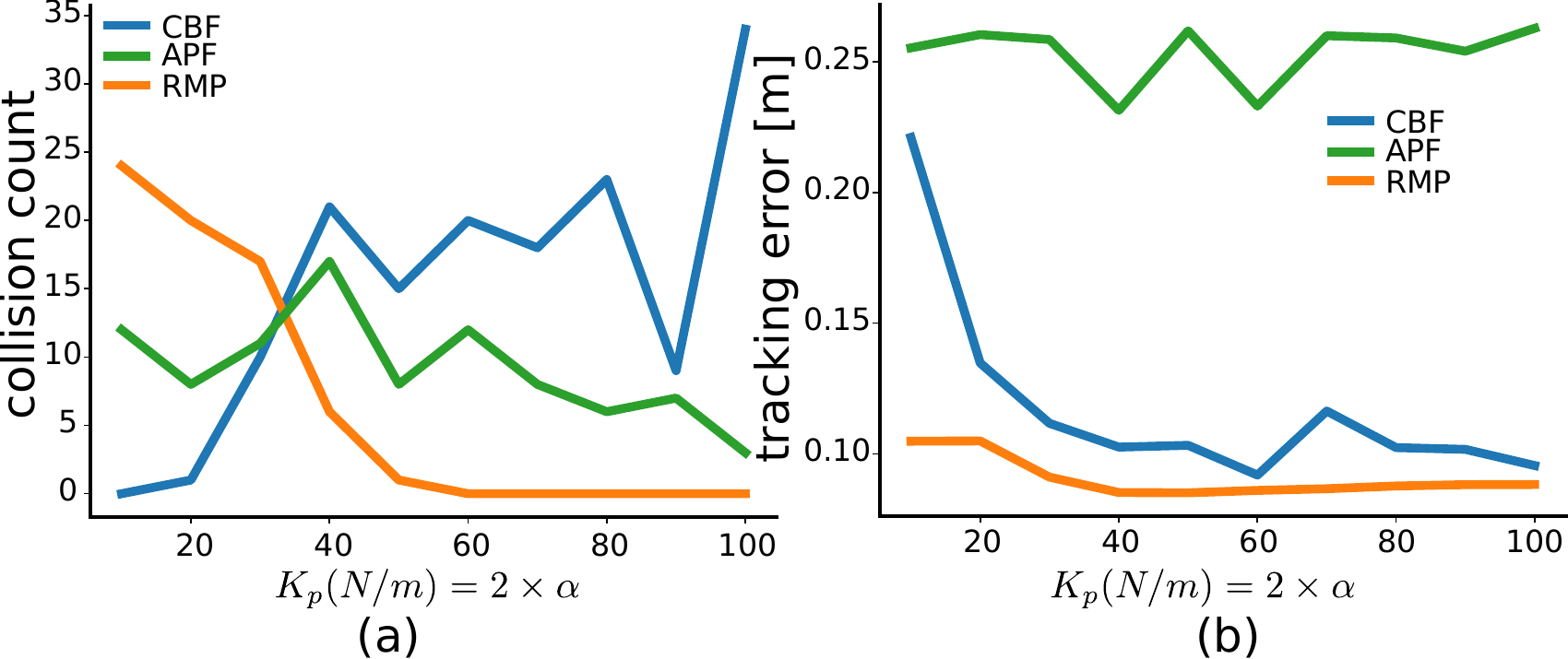}
    \caption{Comparison of high-speed self-collision avoidance test results using APF, CBF \cite{mit_collision}, and our proposed approach. (a) illustrates the variation in the count of collision instances as we adjust the $\alpha$ parameter of CBF and the $K_p$ parameter of APF and RMP. (b) presents the mean total tracking error, measured as the Euclidean distance between the desired limb trajectory and the actual trajectory throughout the entire motion period. This mean value represents the average tracking error across all four limbs.}
    \label{fig:cbf_rmp}
\end{figure}

In order to assess the efficacy of different reactive controller-based methods under 
high-speed operations, we performed self-collision avoidance tests using a simulation of the MIT Mini-Cheetah quadruped robot\cite{mini-cheetah}. During these experiments, the robot was inverted and tasked with tracking limb trajectories that involved numerous self-collision phases while trying to maintain a fixed body position, as illustrated in Figure \ref{fig:mc_exp}.
Each cycle of the maneuver, from (a) to (d) and back in Figure \ref{fig:mc_exp}, lasted 0.05 seconds, and the entire sequence of motion was sustained for five seconds. We evaluated the effectiveness of three distinct approaches: a conventional Artificial Potential Field (APF) method that relies solely on repulsive force, a Control Barrier Function (CBF) optimization approach as elaborated in \cite{mit_collision}, and our Riemannian Motion Policy (RMP)-based strategy. Both the APF and CBF methodologies were implemented in accordance with the procedures outlined in \cite{mit_collision}. To ensure a balanced comparison, we administered the tests with a variety of parameter configurations. Specifically, we assessed different settings of the $\alpha$ parameter in the CBF-based method\cite{mit_collision}. This parameter was subsequently doubled to obtain the value of the proportional gain $K_p$ that was subsequently employed to compute the repulsive force in both the APF and RMP methods.
As illustrated in Figure \ref{fig:cbf_rmp}, the baseline APF-based approach struggles with self-collision avoidance and exhibits poor tracking performance. The CBF method, despite its proficiency at avoiding collisions with low settings of the $\alpha$ parameter, unfortunately, falls short in tracking performance when compared to the RMP-based method. Contrasting these, the RMP-based approach demonstrates an effective balance in maintaining a low tracking error while concurrently mitigating the risk of self-collision, outperforming the aforementioned alternative strategies.
\subsection{Experimental validation of point-foot biped}

A preliminary unassisted in-place stepping test is conducted to validate the viability of the controller and the robot's hardware. In this experiment, the robot was supported by a person for the first few steps and let go afterward. The framework summarized in Fig. \ref{fig:ctrl_framework} is used for the hardware experiments. In addition to the proposed controller, kinematics-based WBC is used to improve foot placement accuracy\cite{wbic, IJRR}. In this experiment, collision avoidance RMP is not used as the experimental platform is not yet mature enough to conduct the aggressive leg crossing movements.

Fig.\ref{fig:hw_result} depicts tracking performance during the stepping test and shows that the controller can follow the commanded floating base references closely. Note that, even though we only do yaw control during the $0.024~\si{\second}$ long dual support phases, Pat can still track the commanded zero yaw. In this work, we achieved up to forty unassisted steps and a video recording of the experiments can be found in the attached supplementary material.


\section{CONCLUSION AND DISCUSSION}
This paper proposes a new formulation for integrating the RMPflow computational framework with a nullspace projection-based whole-body controller. Based on the proposed formulation, a collision-avoidance swing-leg controller was designed and validated in simulation. Moreover, we presented a new small-scale point-foot biped robot purpose-built for experimental studies of dynamic biped locomotion. The hardware experiment results presented in this paper show the robot's viability but also indicate room for improvement. In the future, we plan to replicate the push recovery experiments conducted in simulation on the physical robot.

\section*{Acknowledgement}
This material is based upon work supported by the National Science Foundation under Grant No. 2220924.

\bibliographystyle{IEEEtran}
\bibliography{root}

\end{document}